\documentclass[runningheads]{llncs}
\usepackage{graphicx}
\usepackage{booktabs}
\usepackage{amsmath}
\usepackage{amsfonts}
\usepackage{booktabs}
\usepackage{threeparttable}
\graphicspath{{figures/},{pics}}
\usepackage{subfigure}
\usepackage{booktabs}
\usepackage{multirow}
\usepackage{array}
\usepackage[table]{xcolor}
\usepackage{marvosym}

\begin{document}
\title{Self-Ensembling Contrastive Learning for Semi-Supervised  Medical Image Segmentation}
\titlerunning{Self-Ensembling Contrastive Learning}
%
\author{Jinxi Xiang\inst{1}\thanks{Work done when Jinxi Xiang was an intern at SenseTime Research.}, Zhuowei Li\inst{2}, Wenji Wang\inst{2}, Qing Xia\inst{2}\textsuperscript{\Letter}, Shaoting Zhang\inst{2}}

\authorrunning{J. Xiang et al.}
\institute{Tsinghua University, Bejing, China \and
	SenseTime Research, China \\
	\textsuperscript{\Letter}\email{xiaqing@sensetime.com}\\}

%
\maketitle              
\begin{abstract}
    Deep learning has demonstrated significant improvements in medical image segmentation using a sufficiently large amount of training data with manual labels. Acquiring well-representative labels requires expert knowledge and exhaustive labors. In this paper, we aim to boost the performance of semi-supervised learning for medical image segmentation with limited labels using a self-ensembling contrastive learning technique. 
 	To this end, we propose to train an encoder-decoder network at image-level with small amounts of labeled images, and more importantly, we learn latent representations directly at feature-level by imposing contrastive loss on unlabeled images. This method strengthens intra-class compactness and inter-class separability, so as to get a better pixel classifier. Moreover, we devise a student encoder for online learning and an exponential moving average version of it, called teacher encoder, to improve the performance iteratively in a self-ensembling manner. 
 	To construct contrastive samples with unlabeled images, two sampling strategies that exploit structure similarity across   medical images and utilize pseudo-labels for construction, termed region-aware and anatomical-aware contrastive sampling, are investigated. 
	We conduct extensive experiments on an MRI and a CT segmentation dataset and demonstrate that in a limited label setting, the proposed method achieves state-of-the-art performance. Moreover, the anatomical-aware strategy that prepares contrastive samples on-the-fly using pseudo-labels realizes better contrastive regularization on feature representations.
\end{abstract}
\section{Introduction}
Major successes have been reported using supervised learning for medical segmentation using a large and well-representative dataset\cite{Milletari2016VNetFC} \cite{iek20163DUL}. Manual labeling of medical images by physicians is time-consuming and laborious. Training CNNs only using a small number of labeled images largely decrease the performance of networks. Semi-supervised learning aims to mitigate the strong requirement of data annotation by deeply exploiting unlabeled data information \cite{Chen2019SelfsupervisedLF} \cite{Zhu2020RubiksCA}.

Many deep semi-supervised learning approaches have been introduced, including proxy-label methods \cite{Ruder2018StrongBF}, generative models \cite{Odena2016SemiSupervisedLW}, graph-based methods \cite{Zhu2003SemiSupervisedLU}. A primary line of work is inspired by imposing consistency regularization that favors a more consistent model whose predictions of one example should not change significantly under random perturbations \cite{Tarvainen2017MeanTA} \cite{Ke2019DualSB} \cite{Xie2020SelfTrainingWN}.
Consistency regularization is a \textit{image-level} paradigm that learn latent representations with the loss measured in the output space but it does not exploit inter-/intra-class features. Moreover, the random perturbations are inefficient in high dimensions \cite{Verma2019InterpolationCT}.
Most recently, contrastive learning largely shortens the gaps between supervised and un-/self-supervised learning on visual representations. By discriminating among unlabeled images from different groups, contrastive learning attempts to maximize inter-class differences and to maximize intra-class agreements. The performance of  resulting CNN is greatly enhanced in downstream tasks such as image classification on ImageNet \cite{Chen2020ASF} \cite{He2020MomentumCF}. However, the applicability of contrastive learning in medical image segmentation is yet to be explored  \cite{Iwasawa2020LabelEfficientMS}.

In this paper, we propose to embed the contrastive loss at the feature level to extract feature maps with richer information by transferring the robust visual representations of the contrastive learning mechanism to the medical image segmentation tasks, rather than using the consistency loss at the image level.
Two reasons for superiority: 
(1) Only capturing the pattern information from the image level is not powerful enough to enhance the feature-level representation \cite{Khosla2020SupervisedCL}. Capturing pattern information at the \textit{feature level} is more straightforward to learn latent representations. By explicitly performing a discriminative task between samples, we encourage features of pixels from the same group to be compact and otherwise to be separated. This results in a more robust and accurate final segmentation performance.
(2) The use of strong data augmentation between views, rather than weak random noise perturbation, can be applied to allow contrastive learning to benefit from spatial co-occurrence. The increasing data augmentation can lead to decreasing mutual information between views and improves downstream task performance \cite{Tian2020WhatMF}.

We further address the  problem of constructing positive and negative samples for contrastive loss. For image classification, it is intuitive to do instance discrimination \cite{Wu2018UnsupervisedFL}. To construct 'instances' for segmentation,  it is necessary to integrate domain-specific knowledge \cite{Chaitanya2020ContrastiveLO}. 
This observation motivates us to exploit the structure information of medical images to construct contrastive samples, more concretely, by using the similarity contents between images as clues or using pseudo-labels of unlabeled images to guide anatomical structure. They are termed  region-aware and anatomical-aware strategies, respectively.

In a nutshell, we present a novel self-ensembling contrastive learning architecture for semi-supervised medical image segmentation, combining the \textit{image-level} supervised loss and \textit{feature-level} contrastive loss. Our main contributions are twofold: (1) propose a self-ensembling semi-supervised learning architecture that exploits feature-level contrastive learning representation using unlabeled data; (2) propose region-aware and anatomical-aware strategies for contrastive sample construction in medical image segmentation scenario.

\section{Methods}
%

\subsection{Self-Ensembling Contrastive Learning Framework}


Given $N$ labeled training data and $M$ unlabeled training data, denoted as $\mathcal{D}_{L}=\left\{\left(x_{i}, y_{i}\right)\right\}_{i=1}^{N}$ and $\mathcal{D}_{U}=\left\{\left(x_{i}, y_{i}\right)\right\}_{i=N}^{N+M}$, the problem is formulated to minimize the following  loss:
\begin{equation}
	\min _{\theta} \frac{1}{N} \sum_{x_i \in \mathcal{D}_{L}} \mathcal{L}_{s}\left(g_\theta\left(f_\theta\left(x_{i}\right)\right), y_{i}\right)+\lambda \frac{1}{M}\sum_{x_i \in \mathcal{D}_{U}} \mathcal{L}_{c}\left(\tilde{z}_i, \hat{z}_i\right)
	\label{eq:objective}
\end{equation}
where $\mathcal{L}_s$ denotes the supervised loss (e.g., cross-entropy loss);
$\mathcal{L}_c$ denots the unsupervised contrastive loss for augmented views $\tilde{x}_i, \hat{x}_i$;
the vector representions are extracted as $\tilde{z}_i = p_\theta\left(f_\theta\left( \tilde{x}_i\right)  \right), \hat{z}_i =  p_\zeta\left(f_\zeta\left( \hat{x}_i\right)  \right)$;  
$\lambda$ is a ramp-up coefficient of unsupervised loss: $\lambda(t)=0.1 * e^{\left(-5\left(1-t / t_{\max }\right)^{2}\right)}$. 

The  contrastive loss $\mathcal{L}_c$ at feature-level encourage the encoder to learn good visual representations that discriminate samples from different distributions. The first essential component of contrastive loss is a data augmentation module that transforms a given data randomly into different views, $\hat{x}$ and $\tilde{x}$. The improvements of representations closely related to the quality of augmentation views used for training. We apply five augmentations: random intensity shift, random elastic deformation, random flip, random scale, random rotation. A neural network encoder $f_\theta$ is employed to extract feature maps from augmented data and then a shallow multiple perceptron layer projector is further used to transform the feature maps to lower-dimensional vectors and preserve more information at the output of the encoder. A contrastive loss is considered in the latent space:
\begin{equation}
	\mathcal{L}_c(\tilde{z}_i, \hat{z}_i)=-\log \frac{\exp ( {\operatorname{sim}(\tilde{z}_i, \hat{z}_i)  / \tau})}{\exp ({\operatorname{sim}(\tilde{z}_i, \hat{z}_i) / \tau)}+\sum_{z_l \in \Lambda^{-}} \exp  ({\operatorname{sim}\left(\tilde{z}_i, z_l\right) / \tau)}}
	\label{eq:InfoNCE}
\end{equation}
where $\operatorname{sim}(\boldsymbol{u}, \boldsymbol{v})=\boldsymbol{u}^{\top} \boldsymbol{v} /\|\boldsymbol{u}\|\|\boldsymbol{v}\|$  denotes the cosine similarity; $\tau$ is the temperature used to smooth ($\tau > 1$) or sharpen ($\tau < 1$) the distribution; $\{ \tilde{z}_i, \hat{z}_i \}\in \Lambda^{+}$ is a positive sample pair and  $\{ \tilde{z}_i, z_l\} \in \Lambda^{-}$ is a negative sample pair. 

Ensembling the  encoder at different training stages can further bootstrap the quality of new representations. We use representations from the teacher encoder as targets to guide the student encoder for training the next enhanced representation.
By iterating between the student and teacher encoder,  a series of representations is built with increasing quality by updating the teacher's weights as an exponential moving average (EMA) of the student's weights. In this way, the information from unlabeled data at different training stages is gradually ensembled.  
Given $\alpha\in [0,1]$, the EMA we perform:
\begin{equation}
	\begin{array}{l}
		\boldsymbol{f}_{\zeta}^{(t)} \leftarrow \alpha \boldsymbol{f}_{\zeta}^{(t-1)}+(1-\alpha) \boldsymbol{f}_{\boldsymbol{\theta}}^{(t-1)} \\
		\boldsymbol{p}_{\zeta}^{(t)} \leftarrow \alpha \boldsymbol{p}_{\zeta}^{(t-1)}+(1-\alpha) \boldsymbol{p}_{\boldsymbol{\theta}}^{(t-1)}
	\end{array}
\label{eq:ema}
\end{equation}

Based on the representation of the student encoder, we adopt a decoder to recursively use features with high-level semantic information (low-resolution image) to refine the features of low-level information (high-resolution image), until we get feature maps with the same resolution as the input image. Supervised loss is computed in the output space with labels.

\subsection{Leveraging Structure Information for Contrastive Loss}

Two strategies are investigated to construct positive and negative samples for contrastive loss leveraging structure information.

\textbf{Region-Aware Contarstive Sampling (RACS).} 
We assume that   images of the same anatomical region from  different subjects have similar content if they are roughly aligned. 
We exploit structure similarities across corresponding regions in different subjects as additional knowledge for constructing positive and negative samples in the contrastive loss \cite{Chaitanya2020ContrastiveLO}. 
As shown in Fig. \ref{fig:strategy} (a), $\Lambda^{+}$ contains positive pairs of corresponding partitions $s$ across subjects $i,j$ and their transformed versions, that is,  $\left(\tilde{x}_{i}^{s}, \tilde{x}_{j}^{s}, \hat{x}_{i}^{s}, \hat{x}_{j}^{s}\right)\in\Lambda^{+}$. $\Lambda^{-}$ contains the remaining patitions and their transformations. 
One prerequisite of the method is that the input images must be roughly aligned. This requirement can be met when the images are acquired with the same modality and set to capture the same field-of-view. Otherwise, it might cause detrimental results.

\textbf{Anatomical-Aware Contrastive Sampling (AACS).} 
In semi-supervised framework, we can first employ the existing encoder-decoder model $(f_\theta, g_\theta)$ trained on limited labeled data to predict pseudo-labels for each data point from the unlabeled set, and then utilize the pseudo-labels to assist contrastive samples construction, as illustrated in Fig. \ref{fig:strategy} (b). Specifically, pseudo-labels are used to locate the \textit{anatomical centers} of each substructure in the unlabeled images, and then a minibatch of cubes $\{x_1, x_2, ...x_K \}$ is cropped from the unlabeled   images. To cover all substructures of the input images under consideration, $K\geq C$ where $C$ is the number of classes in the segmentation task. Two augmentation views are constructed from this minibatch, resulting in 1 pair of positive samples and the remaining $2(K-1)$ are negative samples.
At the initial stages of training, the resulting pseudo-labels from $(f_\theta, g_\theta)$ are of low quality. The ramp-up weight $\lambda$ of contrastive loss is small at this phase. As the network progressively improved, the pseudo-labels are getting better which in return boosts the network training.


\section{Experiments and Results}

We have conducted two sets of experiments on public datasets to evaluate the proposed self-ensembling contrastive learning for semi-supervised   medical image segmentation. The first set of experiments is conducted on MRI ACDC-2017 with 3 substructures (LV, RV, MYO) \cite{Bernard2018DeepLT} whereas the second on MMWHS-2017 CT with 7 substructures (LV, RV, LA, RA, PA, AO, MYO) \cite{Zhuang2019EvaluationOA}. \textit{Results on ACDC-2017 can be found in the supplementary material.}
We evaluate one fully-supervised learning approach, i.e.   U-Net, three semi-supervised learning approaches, i.e. (1) standard mean teacher \cite{Tarvainen2017MeanTA}, (2) mean teacher with uncertainty aware \cite{Yu2019UncertaintyawareSM}, and (3) noisy student \cite{Xie2020SelfTrainingWN}, along with our proposed SECL using two contrastive sampling strategies. All the above-mentioned approaches employ  3D-UNet as the backbone network. The network is trained using ADAM optimizer with the learning rate set to $10^{-3}$ for 1000 epochs.

\newcolumntype{C}[1]{>{\centering\let\newline\\\arraybackslash\hspace{0pt}}m{#1}}

\textbf{Whole Heart Segmentation Dataset \cite{Zhuang2019EvaluationOA}.} In total, this dataset provided 120 multi-modality whole heart images from multiple sites, including 60 cardiac CT. We selected the CT images and split 20 images to form the training set, and the remaining 40 to form the test set. We partition the 20 labeled data into 10 training and 10 validation. More importantly, we add up with 50 unlabeled cardiac  CT data collected from centers in order to validate different semi-supervised learning.
We employed three widely used metrics, Dice score,  Jaccard, and Hausdorff Distance (HD) to evaluate the segmentation results on 40 testing data.  The evaluation tool is provided officially by \cite{Zhuang2019EvaluationOA}.

Table \ref{table:whs} shows the results on MMWHS-2017 dataset. Dice and Jaccard of WHS is the weighted average of all seven substructures.  HD is the maximum of mean HD of 40 testing subjects.
In general, semi-supervised approaches outperform the baseline   U-Net method in all three metrics.  The proposed SECL with anatomical-aware sampling leads to the best performance whereas SECL with region-aware sampling has slightly degraded results. With the help of contrastive learning at the feature-level, the performance of   U-Net using small labeled training sets can be improved. More importantly, the construction strategy of contrastive samples has a substantial effect on the improvement of semi-supervised learning. Anatomical contrastive sampling using pseudo-labels to crop   cubes extracts more distinctive local features of each local substructure, and more importantly, without prior requirements of image alignment, which might not be satisfied in many scenarios.

\begin{table}[tb]
	\caption{Methods Evaluations on Whole Heart Segmentation}
	\label{table:whs}
	\centering
	\begin{threeparttable}
		\begin{tabular}{C{4mm}C{30mm}C{10mm}C{10mm}C{10mm}C{10mm}C{10mm}C{10mm}C{10mm}C{11mm}}
			\toprule[0.02in]
			 &  \textbf{Methods} & LV & RV & LA & RA & MYO  & AO & PA & \textbf{WHS}\\ \midrule[0.01in]
			\multirow{6}{*}{\rotatebox[origin=c]{90}{\textbf{Dice}}} &   U-Net \cite{iek20163DUL} &  0.8806 &	0.8386 & 0.8375 &	0.8789 &	0.7849 &	0.8054 &	0.7068 &	0.8435       \\
			& Mean Teacher \cite{Tarvainen2017MeanTA}  & 0.8791	& 0.8432	& 0.8477	& 0.8857	& 0.8351	& 0.8604	& 0.7903	& 0.8609 \\
			& MT-UA \cite{Yu2019UncertaintyawareSM} &  0.8914 &	0.8478 &	0.8470 & 	0.8855 &	0.8259 &	0.9085 &	0.7623 &	0.8639			      \\   
			& Noisy Student \cite{Xie2020SelfTrainingWN} &    0.8946 &	0.8412 &	0.8460 &	0.8845 &	0.8217 &	0.8720 &	0.7218 &	0.8587			   \\ 
			&  \textbf{SECL (RACS)} & 0.8978 & 0.8763 &	0.8721 &	0.8883 &	0.8474 &	0.8684 &	0.7893 &	0.8765 					 \\ 
            \rowcolor{blue!5}	&  \textbf{SECL (AACS)} &0.9138 &	0.8573 &	0.8815 &	0.9111 &	0.8226 &	0.9294 &	0.7837 &	{\textbf{0.8843}}        \\ \midrule
			\multirow{6}{*}{\rotatebox[origin=c]{90}{\textbf{Jaccard}}} &   U-Net \cite{iek20163DUL}&  0.8144 &	0.7439 &	0.7416 &	0.7984 &	0.6765 &	0.7362 &	0.5990 &	0.7489 \\
			& Mean Teacher \cite{Tarvainen2017MeanTA} & 0.8010 &	0.7362 & 	0.7485 &	0.8035 &	0.7303 &	0.7821 &	0.6797 &	0.7626			       \\
			& MT-UA  \cite{Yu2019UncertaintyawareSM} & 0.8208 &	0.7488 &	0.7522 &	0.8044 & 0.7164 &	0.8379 &	0.6428 &	0.7689 			       \\   
			& Noisy Student \cite{Xie2020SelfTrainingWN} &   0.8164 &	0.7306 &	0.7395 &	0.8002 &	0.7110 &	0.7950 &	0.6122 &	0.7570 			     \\ 
			& \textbf{SECL (RACS)} & 	0.8269 &	0.7832 &	0.7781 &	0.8070 &	0.7440 &	0.8029 &	0.6829 &	0.7854 \\ 
			\rowcolor{blue!5} & \textbf{SECL (AACS)} & 0.8535 &	0.7708 &	0.7716 &	0.8402 &	0.7208 &	0.8667 &	0.6583 &	\textbf{0.7927}					\\ \midrule
			\multirow{6}{*}{\rotatebox[origin=c]{90}{\textbf{HD ($mm$)}}} &   U-Net \cite{iek20163DUL} &  31.58 &	14.97 &	8.356 & 14.27	 &	64.23 &	8.307 & 23.59	 &	 18.43		\\
			& Mean Teacher  \cite{Tarvainen2017MeanTA} & 6.467 &	4.880 & 4.131	 & 21.29	 &	5.875 &	 7.408 & 9.695	 &		5.546		       \\
			& MT-UA   \cite{Yu2019UncertaintyawareSM} & 3.441 &	 5.526 & 4.842	 &	10.281 & 6.812 & 3.673	 &	19.42 &	 7.407			       \\   
			& Noisy Student \cite{Xie2020SelfTrainingWN} & 6.766    &  7.387      &  10.72      & 7.742     &  2.741       & 5.491   & 24.08 &  6.297   \\ 
			& \textbf{SECL (RACS)} & 6.011 & 6.944 & 5.547 & 7.062   & 3.274  &  9.133 & 8.288  &	5.617		 \\ 
			\rowcolor{blue!5} & \textbf{SECL (AACS)} & 3.708 &	 3.922 & 2.618	 & 7.692	 & 2.076	 &	2.032 &	15.09 &	\textbf{3.221}			\\ \bottomrule[0.02in]
		\end{tabular}
	\end{threeparttable}
\end{table}

\section{Ablation Study}

We present ablations on SECL to give an intuition of its behavior and performance.  Two important aspects of SECL are studied: (1) the self-ensembling of the teacher-student encoder for contrastive representation learning. Whether using a single model ($\alpha=0$ in Eq. (\ref{eq:ema})) rather than teacher-student structure achieve similar results? (2)  the intuitive and essential results of imposing contrastive loss on encoder at feature-level.  Ablation studies are made on the MMWHS-2017 dataset.

\subsection{Exploring Contrastive Self-Ensembling}

SECL uses the vector representations of the teacher model as targets for the predictions of the student model. Since the weights of the teacher model are EMA of the student model, the teacher model is considered to be a delayed and more stable version of the weights of the student network. At different training processes, the student model attempts to approximate the output of the teacher model through online learning, and thus iteratively achieves self-ensembling. 
As shown in Table \ref{table:ema}, letting $\alpha=0$, we get a single encoder model which has similar structure with multi-task learning with contrastive regularization branch in \cite{Iwasawa2020LabelEfficientMS}. Without self-ensembling, it did not stabilize the training process, leading to poor performance.  On the other hand, overly large momentum update factors i.e. $\alpha=0.99$  results in slow update of the teacher model and prevents final improvements. We consider  $\alpha=0.9$ is an appropriate value in this setting.  

\begin{table}[!ht]
	\caption{Self-Ensembling of Teacher-Student Encoder (Dice Score)}
	\label{table:ema}
	\centering
	\begin{threeparttable}
		\begin{tabular}{C{28mm}C{10mm}C{10mm}C{10mm}C{10mm}C{10mm}C{10mm}C{10mm}C{12mm}}
			\toprule[0.02in]
			\textbf{Methods} & LV & RV & LA & RA & MYO  & AO & PA & \textbf{WHS}\\ \midrule[0.01in]
			SECL ($\alpha=0$)   & 0.8999 &	0.8483 &	0.8362 &	0.8789 &	0.8346 &	0.8661 &	0.6689 &	\textbf{0.8562} \\
			SECL ($\alpha=0.8$)   &  0.9033 &	0.8523 &	0.8693 &	0.9159 &	0.7881 &	0.9031 &	0.8096 &\textbf{0.8734} \\
			 \rowcolor{blue!5} SECL ($\alpha=0.9$)  & 0.9138 &	0.8573 &	0.8815 &	0.9111 &	0.8226 &	0.9294 &	0.7837 &	\normalsize{\textbf{0.8843}} \\  
			SECL ($\alpha=0.99$)  & 0.9074 &	0.8624 &	0.8759 &	0.9160 &	0.8349 &	0.9174 &	0.7299 &	\textbf{0.8793} \\ \bottomrule[0.02in]
		\end{tabular}
	\end{threeparttable}
\end{table}

\subsection{Building Intuitions on Contrastive Loss}
SECL is underpinned by contrastive representation learning aiming to discriminate among different class patches at the feature-level. We create a batch of substructure samples with anatomical-aware sampling from 20 training data and then apply five data augmentations on it.
The augmented data are used as the input to compare the vector representation from the trained encoder of the Mean Teacher and from the proposed SECL with anatomical-ware sampling.  
We reduce the vector dimension to 2D coordinates and visualize the embedding vectors  with t-SNE toolkits \cite{Maaten2008VisualizingDU}.

Fig. \ref{fig:WHS} shows the distribution of seven classes of substructure on the whole heart segmentation dataset using the standard mean teacher model and the proposed SECL (AACS) from the output of encoders.
The clustering results of the Mean Teacher model are ambiguous. In comparison, the class support regions are more compact within classes and more separated between classes when trained with the proposed strategy, leading to better final segmentation performance. 
The distribution also agrees with the physical reasons. For example, MYO is much close to LV; AO is most separable in the anatomical sense, etc.


\section{Conclusion}
In this paper, we propose to leverage the unlabeled images by imposing contrastive loss at the feature-level, along with the supervised loss at the image-level. The self-ensembling mechanism further enhances the quality of representation. Two contrastive sampling strategies for contrastive loss are presented and the anatomical-aware sampling show superior performance in exploiting refined local information over the region-aware sampling. The contrastive loss could be a complimentary regularization to existing semi-supervised models and we suppose that their combination may yield better accuracy than either of them alone.


%

\end{document}